\documentclass[letterpaper]{article} % DO NOT CHANGE THIS
\usepackage{aaai2027}  % DO NOT CHANGE THIS
\usepackage[hyphens]{url}  % DO NOT CHANGE THIS
\usepackage{graphicx} % DO NOT CHANGE THIS
\usepackage{natbib}  % DO NOT CHANGE THIS AND DO NOT ADD ANY OPTIONS TO IT
\usepackage{caption} % DO NOT CHANGE THIS AND DO NOT ADD ANY OPTIONS TO IT
\usepackage{algorithm}
\usepackage{algorithmic}
\usepackage{subcaption}

\usepackage{newfloat}
\usepackage{makecell}
\usepackage{listings}
\DeclareCaptionStyle{ruled}{labelfont=normalfont,labelsep=colon,strut=off} % DO NOT CHANGE THIS
\floatstyle{ruled}
\newfloat{listing}{tb}{lst}{}
\floatname{listing}{Listing}

\newcommand{\eg}{\textit{e.g.}}
\newcommand{\ie}{\textit{i.e.}}
\newcommand{\etal}{\textit{et al.}}

\makeatother

\usepackage{booktabs}

\nocopyright

\usepackage{multirow}
\usepackage{amsmath}
\usepackage{amssymb}
\usepackage{xcolor}
\usepackage{pifont}
\usepackage{array}

\definecolor{darkred}{RGB}{200,0,0}
\definecolor{darkgreen}{RGB}{0,200,0}

\usepackage{hyperref}

\definecolor{myorange}{RGB}{245,156,74}
\definecolor{mygray}{gray}{0.4}
\definecolor{myred}{RGB}{197,41,114}

\hypersetup{
  colorlinks=true,
  urlcolor=blue,
  linkcolor=red,
  citecolor=-myorange,
}

\newcommand{\yesmark}{\textcolor{darkgreen}{\ding{51}}}
\newcommand{\nomark}{\textcolor{darkred}{\ding{55}}}

\title{Improving Complex Moir\'e Removal with Generative Supervision}

\author{
    Xinyang Gu\textsuperscript{\rm 1},
    Zhilu Zhang\textsuperscript{\rm 1},
    Honglei Xu\textsuperscript{\rm 1},
    Yanting Mei\textsuperscript{\rm 1},
    Yukang Ding\textsuperscript{\rm 2},
    Wangmeng Zuo\textsuperscript{\rm 1}
}

\affiliations{
    \textsuperscript{\rm 1}Harbin Institute of Technology, Harbin, China\\
    \textsuperscript{\rm 2}Alibaba Group - Taobao \& Tmall Group\\
    \vspace{2mm}
}



\begin{document}
\pagestyle{plain}
\setcounter{page}{1}

\maketitle

\begin{abstract}

The availability of high-quality paired data is essential for training learning-based image demoir\'eing models. However, it remains challenging for existing datasets to encompass the complex moir\'e patterns captured in uncontrolled real-world scenarios. Such degradations typically manifest as large-scale, multicolored moiré patterns. Moreover, these patterns frequently occur in images for which clean counterparts are difficult to obtain, such as photographs acquired from public displays or existing online resources. In this work, we propose a novel data engine designed to improve the removal of complex moir\'e patterns by generating training supervision. Specifically, we initially collect real-world images containing complex moir\'e patterns and localize the corresponding screen regions. Multiple image-conditioned generative foundation models are subsequently deployed to produce candidate references. To establish reliable supervision, these candidates are subjected to patch-level quality control to filter and select the optimal results. Based on this systematic paradigm, we construct the WildMoir\'e dataset, which contains 6.8K moir\'e-GT training pairs. For evaluation, we additionally build an independent test set comprising $\sim$250 pairs with captured clean ground truth. Extensive experiments on ESDNet, SDXL, and Qwen-Image-Edit demonstrate that the proposed generative supervision consistently improves the performance of complex moir\'e removal. Project: \url{https://xinygu-pavo.github.io/WildMoire/}.

\end{abstract}

\section{Introduction}
Using smartphones and digital cameras to record information from electronic screens has become increasingly common in daily life. However, captured screen images often
contain moir\'e artifacts that are absent from the original displayed content. These
artifacts may appear as fine colored grids, curved waves, broad low-frequency color
bands, and mixtures of several patterns. They arise because
both the display and the camera sensor sample visual signals on discrete grids, and
interference between the display pixel grid and the camera sensor array can produce
objectionable aliasing patterns~\cite{liu2018demoir}. Moir\'e artifacts not only
degrade visual quality, but may also obscure text, alter colors, and destroy image
textures and structural details. Therefore, removing moir\'e from camera-captured
screen images is an important problem for mobile photography, digital-signage
capture, document recording, and visual-content sharing.

\begin{figure}[t]
\centering
\centering
    \includegraphics[width=0.48\textwidth]{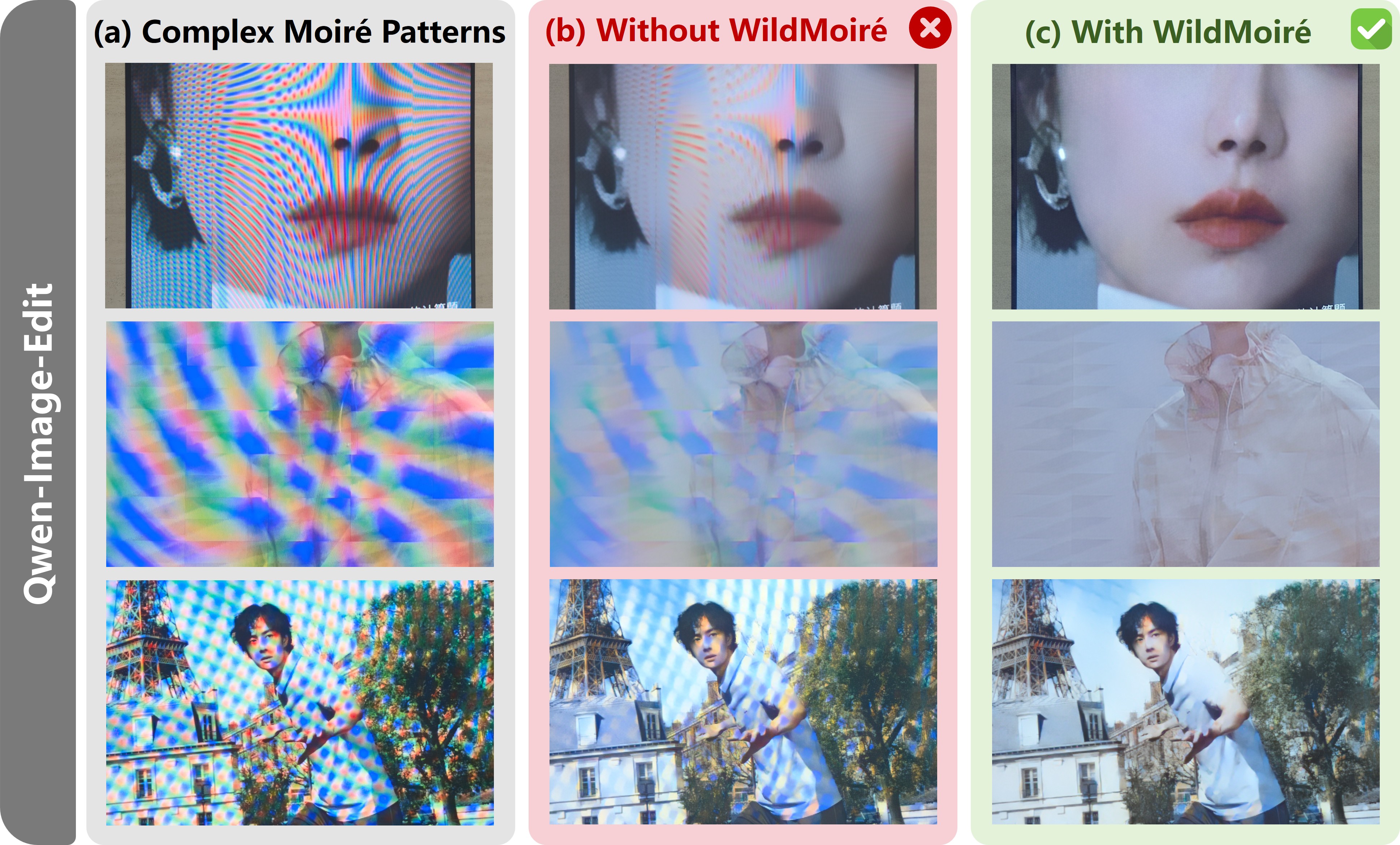}
\caption{Results on moir\'e images. The (a) column presents three moir\'e inputs with large-scale chromatic bands and overlapping interference patterns. The (b) and (c) columns show Qwen-Image-Edit trained without and with our constructed WildMoir\'e dataset, respectively. Training only on the previous dataset leaves artifacts and suppresses details, while incorporating WildMoir\'e removes the broad interference and preserves original contents better.}
\label{fig:motivation}
\end{figure}

Facilitated by the collection of large-scale paired data~\cite{sun2018dmcnn,he2020fhde2net,yu2022esdnet,mei2025image}, learning-based image demoir\'eing methods have achieved substantial progress through multi-scale processing strategies~\cite{zheng2020mbcnn,cheng2023recaptured}, the integration of attention mechanisms~\cite{he2025universal}, and the introduction of diffusion models~\cite{zhu2026combined,yang2026realrestorer}. Nevertheless, these trained demoir\'eing models yield satisfactory results only within the distribution of the training data, exhibiting limited generalizability across different screens and complex scenes, especially for large-scale, multicolored moir\'e patterns (see Fig.~\ref{fig:motivation}(b)). A straightforward approach is to continuously collect and scale up paired data for training. However, this solution is impractical and expensive in terms of both manpower and computational resources, owing to the diverse range of existing displays and the continuous emergence of new devices.

Fortunately, recent foundation models for image generation and editing, \eg, Nano-Banana-2~\cite{google2026nanobanana2} and GPT-Image-2~\cite{openai2026gptimage2}, demonstrate remarkable generalization capabilities for moir\'e removal when provided with carefully crafted prompts, which stems from their rich and strong visual priors. But simultaneously, these models are difficult to adopt directly as professional-grade demoir\'eing tools because they are prone to generating hallucinations, \eg, inconsistencies in content and color. To address this issue, we propose to post-process and filter these outputs to serve as supervision for demoir\'eing models, which functions as a more cost-effective and convenient solution for scaling up paired data. Moreover, instead of relying on the outputs of a single generative model, we compare the results from multiple generative models and select the optimal output for supervision. This strategy aims to leverage the complementary strengths across different models.

Specifically, we initially collect a substantial number of complex moir\'e images through on-site photography and existing online resources. Subsequently, we localize the screen regions and process them using foundation models for image generation and editing. Given that different models exhibit distinct trade-offs between moir\'e removal and content preservation, we deploy multiple models. These include closed-source models (\ie, GPT-Image-2~\cite{openai2026gptimage2} and Nano-Banana-2~\cite{google2026nanobanana2}) and open-source models (FLUX.2~\cite{bfl2025flux2}, SDXL~\cite{podell2023sdxl}, and Qwen-Image-Edit~\cite{wu2025qwenimage}) that have been fine-tuned on existing demoir\'eing datasets. The corresponding outputs are aligned with the input images in terms of spatial positions and colors, and are subjected to patch-based quality control. Finally, the optimal output among them is selected as the ground truth (GT). In total, we construct 6.8K moir\'e-GT pairs with a resolution of 1024$\times$1024, forming the WildMoir\'e dataset. 
Furthermore, we introduce a scale transformation strategy for data augmentation. By rescaling the moir\'e images, the scale of the moir\'e patterns is correspondingly altered, thereby further enhancing the generalization capability of demoir\'eing models.

We additionally collect $\sim$250 pairs with captured clean ground truth images for the evaluation of complex moir\'e removal. Utilizing the proposed WildMoir\'e dataset and data augmentation strategy, we fine-tune the CNN-based ESDNet~\cite{yu2022esdnet}, the diffusion-UNet-based SDXL~\cite{podell2023sdxl}, and the diffusion-Transformer-based Qwen-Image-Edit~\cite{wu2025qwenimage} for experimental validation. The results demonstrate significant improvements in both quantitative metrics and visual quality. Moreover, comprehensive ablation studies confirm the effectiveness of the various components in the proposed pipeline.

Our contributions are summarized as follows:

\begin{itemize}

    \item We propose generating supervision to extend the training data for the improvement of complex moir\'e removal. Specifically, we introduce a systematic paradigm that filters and selects the optimal outputs from multiple generative models to serve as high-quality supervision.
        
    \item Using the proposed data engine, we construct WildMoir\'e dataset, which contains 6.8K moir\'e-GT pairs. Furthermore, we collect $\sim$250 pairs with captured clean ground truth as an evaluation benchmark.
        
    \item Extensive experiments on ESDNet, SDXL, and Qwen-Image-Edit demonstrate that WildMoir\'e dataset consistently improves the demoir\'eing performance.

\end{itemize}

\section{Related Work}

\subsection{Image Demoir\'eing}

Early image demoir\'eing methods relied on handcrafted signal
decomposition and frequency-domain filtering. With the development
of deep learning, convolutional networks have become the dominant
solution for camera-captured screen images. DMCNN introduced a
multi-resolution architecture to account for the large variation in
moir\'e scales~\cite{sun2018dmcnn}. MopNet classified moir\'e
patterns and processed different categories in a divide-and-conquer
manner~\cite{he2019mopnet}, while MBCNN explicitly modeled
frequency-selective decomposition using learnable bandpass
filters~\cite{zheng2020mbcnn}. To support high-resolution
restoration, FHDe2Net employed a coarse-to-fine framework for
full-HD images~\cite{he2020fhde2net}, and ESDNet introduced
semantic-aligned scale-aware processing for efficient 4K
demoir\'eing~\cite{yu2022esdnet}. More recently, DCID explored dual-camera fusion for mobile image demoir\'eing, using complementary wide- and ultra-wide-camera observations to remove severe moir\'e while preserving high-resolution image details~\cite{mei2025image}.These approaches have substantially advanced model architecture and computational efficiency. Nevertheless, their restoration capability remains closely related to the moir\'e pattern distribution in the training data.

\subsection{Demoir\'eing Data Construction}

Real paired datasets have been essential to the development of
learning-based demoir\'eing. Sun~\etal~ introduced an early benchmark
of real camera-captured screen images~\cite{sun2018dmcnn}.
FHDMi extended paired acquisition to full-HD
resolution~\cite{he2020fhde2net}, while UHDM collected 5,000
ultra-high-definition pairs using multiple smartphones, displays,
and shooting configurations~\cite{yu2022esdnet}. DCID further
constructed 8,959 samples containing wide-angle images,
ultra-wide-angle observations, and digital source supervisions, with
particular attention to severe moir\'e patterns
~\cite{mei2025image}. Although these datasets provide reliable supervision, their acquisition generally requires displaying known clean images on accessible screens and carefully aligning the captured images with their digital sources. Such controlled pipelines are costly to scale and cannot exhaustively cover complex moir\'e patterns encountered in uncontrolled environments.

To alleviate the dependence on costly paired acquisition, another
line of research synthesizes moir\'e-degraded inputs from clean
images. Early work simulated the camera--display imaging process
to construct synthetic training pairs~\cite{liu2018demoir},
while LCDMoir\'e adopted a handcrafted synthesis process for the
AIM image demoir\'eing challenge~\cite{yuan2019aim}. Cyclic
moir\'e learning jointly trained a moir\'eing network and a
demoir\'eing network using unpaired clean and moir\'e
images~\cite{park2022cyclic}. UnDeM divided real moir\'e images
into patches according to their complexity and learned to synthesize
diverse pseudo-moir\'e images for supervised demoir\'eing
training~\cite{zhong2024undem}. More recently, UniDemoir\'e
collected background-independent real moir\'e patterns, employed a
diffusion model to generate additional pattern variations, and
developed a learnable synthesis module to reproduce the color and
brightness characteristics of captured moir\'e images
~\cite{yang2025unidemoire}. These methods improve data scalability by generating the degraded side of a training pair. 

\subsection{Generative Supervision for Image Restoration}

Recent studies have explored generative models as offline producers
of training supervision rather than only as restoration models used
at inference time. HGGT generated multiple enhanced high-resolution
targets for image super-resolution and employed human annotations
to identify regions with beneficial or harmful perceptual changes
~\cite{chen2023hggt}. GMD used a frozen generative oracle to refine
restoration results from unlabeled target-domain images and mixed
the resulting pseudo-pairs with reliable source-domain supervision
for model adaptation~\cite{hu2025gmd}. Additionally, we take note of the concurrent work GGT-100K, which systematically evaluates multimodal foundation models and implements multi-stage quality control to construct a large-scale dataset for general real-world image restoration~\cite{kong2026ggt}. These studies demonstrate the potential of generative supervision, while also highlighting the risks of hallucination, content drift, and inconsistent restoration quality. In this work, we utilize multiple generative models and implement carefully designed quality control measures to construct more reliable supervision signals.

Our work specializes this paradigm for complex screen demoiréing. Instead of relying on a single generator, we use five generative priors with complementary restoration behaviors. Their outputs are spatially and photometrically aligned with the real moiré inputs and evaluated locally before being retained as training supervisions. This design preserves authentic complex moiré degradations on the input side, reduces the influence of unreliable generated content, and produces reusable training data without modifying the deployed restoration architectures or adding test-time generative computation.

\section{WildMoir\'e Dataset}

\begin{figure*}[t]

\centering

\centering
    \includegraphics[width=1\linewidth]{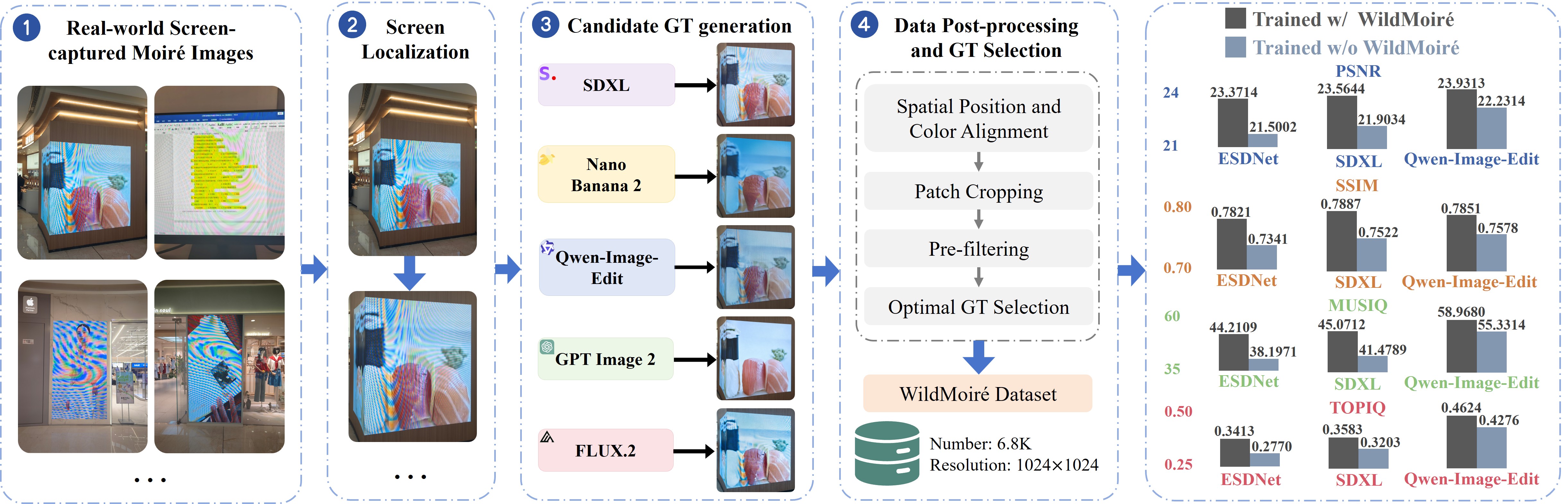}

\caption{Overview of WildMoir\'e dataset construction. We first collect real screen-captured images containing complex moir\'e and localize the display regions. Five image-conditioned generative models then produce candidate GT images. The candidates are processed at matched spatial locations through spatial and color alignment, synchronized patch extraction, pre-filtering, and optimal GT selection. This offline procedure yields 6.8K moir\'e-GT pairs at 1024\(\times\)1024 resolution. The plots on the right summarize the improvements obtained by training ESDNet, SDXL, and Qwen-Image-Edit with constructed WildMoir\'e dataset.}
\label{fig:pipeline}
\end{figure*}

\subsection{Motivation and Problems}

Moir\'e patterns arise from interference between the sensor array of a camera and the pixel grid of a displayed screen.
Existing moir\'e datasets primarily cover simple and small-scale patterns. This limited coverage makes demoir\'eing models less effective in handling complex, large-scale, and multi-color moir\'e patterns, which are particularly common on large public screens.

In fact, obtaining corresponding clean images for such complex moir\'e images is challenging. Firstly, large public displays often present dynamic videos or frequently changing content, making it difficult to recover the exact displayed frame after the moir\'e image has been captured. Secondly, for images collected from online sources, the original clean content is generally unavailable. Fortunately, recent image generation and editing foundation models~\cite{openai2026gptimage2,google2026nanobanana2,bfl2025flux2,podell2023sdxl,wu2025qwenimage} are developing rapidly, and their powerful generative capabilities can provide effective supervision for improving the removal of complex moir\'e patterns.

Although these image generation models exhibit strong restoration capabilities, no single model consistently performs well on all complex moir\'e images. These models show complementarity in moiré removal and content preservation: some effectively remove moir\'e patterns but redraw the image, reducing fidelity, while others retain the original content but leave residual moir\'e artifacts. Therefore, to avoid the limitations of single generative model, a selection strategy is required to identify the highest-quality output as the supervision. By solving these problems, we propose WildMoiré, a dataset designed to improve demoiréing models' ability to remove complex moiré patterns. Its construction pipeline is illustrated in Fig.~\ref{fig:pipeline}.

\subsection{Source Images Collection}
To support robust demoir\'eing of complex moir\'e images, we collect images from two sources. First, we capture 4K-resolution images across diverse scenes using mobile phones. Second, we collect additional images through web crawling to increase diversity of displayed content, screen types, and moir\'e patterns. We intentionally retained challenging examples with broad chromatic bands, overlapping waves, fine repetitive interference, oblique viewing angles, text-rich content, and human subjects, as shown in \textbf{Step~1} in Fig.~\ref{fig:pipeline}. 
After collection, we used Sa2VA~\cite{yuan2025sa2va} to detect screen boundaries and crop the corresponding screen regions, thereby excluding areas that do not contain moir\'e patterns, as shown in \textbf{Step~2} in Fig.~\ref{fig:pipeline}. 

\subsection{Candidate GT Generation}
To obtain corresponding clean supervisions for the collected complex
moir\'e images, we employ SDXL~\cite{podell2023sdxl},
Nano-Banana-2~\cite{google2026nanobanana2},
Qwen-Image-Edit~\cite{wu2025qwenimage},
GPT-Image-2~\cite{openai2026gptimage2}, and
FLUX.2~\cite{bfl2025flux2} to generate candidate GTs, as shown in
\textbf{Step~3} of Fig.~\ref{fig:pipeline}.

As discussed above, different image generation and editing models
exhibit complementary behaviors in restoration capability and
content fidelity. The five selected models provide representative
examples of such complementarity. GPT-Image-2~\cite{openai2026gptimage2} is a powerful
commercial model with strong semantic understanding and editing
capability. It can remove severe moir\'e patterns effectively, but
its aggressive editing may introduce noticeable changes in image
content and color, as well as occasional local hallucinations.
Nano-Banana-2~\cite{google2026nanobanana2} is also a closed-source model and generally
preserves the original content more faithfully than GPT-Image-2~\cite{openai2026gptimage2}.
Nevertheless, it may still introduce content inconsistencies and
leave conspicuous residual moir\'e in some challenging cases.

In comparison, SDXL~\cite{podell2023sdxl}, FLUX.2~\cite{bfl2025flux2}, and Qwen-Image-Edit~\cite{wu2025qwenimage} are open-source
diffusion-based models whose generation behaviors are generally more
conservative after fine-tuning with previous demoir\'eing datasets~\cite{yu2022esdnet,mei2025image}. Although their moir\'e-removal
capability may be weaker than that of the commercial models, they
often preserve the original structures and colors more faithfully.
They therefore provide useful complementary candidates, especially
in regions where GPT-Image-2~\cite{openai2026gptimage2} produces severe hallucinations and
Nano-Banana-2~\cite{google2026nanobanana2} fails to remove the moir\'e artifacts. 
Specifically, to produce reasonable results of moiré removal, SDXL is fine-tuned for 80 epochs using the InstructPix2Pix training framework~\cite{brooks2023instructpix2pix}, while FLUX.2 and
Qwen-Image-Edit are fine-tuned for 100 epochs using
DiffSynth-Studio. At the same time, we also apply the proposed Scale Transformation Data Augmentation strategy, which will be introduced later, to enhance the model's generalization ability.

For candidate generation, the prompts are adapted to each model while preserving a consistent intent: \eg, removing moir\'e and abnormal color interference while retaining text, identities, objects, layout, geometry, and the overall appearance of the photographed display.

\subsection{Post-processing and GT Selection}
\paragraph{Spatial and Photometric Alignment.} 
During candidate GT generation, we observe that each input and its candidate GTs may exhibit spatial and photometric misalignments. To address these issues, we first correct the spatial misalignments and subsequently perform photometric alignment.
Specifically, for spatial alignment, we first employ GlueStick~\cite{pautrat2023gluestick} to estimate a robust global transformation for initial alignment. We then use FlowFormer~\cite{huang2022flowformer} to estimate the dense optical flow between each initially aligned candidate GT and its input for refining the alignment. Finally, we center-crop each input and its aligned candidate GT to remove the misaligned areas around the edges.
For photometric alignment, we first apply Gaussian blurring to the input and its aligned candidate GTs to suppress the influence of high-frequency moir\'e patterns during photometric alignment. We then fit a linear 3\(\times\)3 RGB transformation for each candidate GT to align its color with the input.

\paragraph{Patch Cropping and Pre-filtering.}
This step aims to retain collected input patches with complex moiré and remove clean patches.
We treat each input and its aligned candidate GTs as an image group. For each image group, we sample $n$ spatial locations and crop \(1024 \times 1024\) patches from every image at each location, yielding $n$ patch groups per image group.
Then, we calculate the texture complexity of the input patch in
each patch group using the image-domain standard deviation $\sigma_{img}$~\cite{moulden1990luminance}:
\begin{equation}
\sigma_{img} =
\sqrt{
\frac{1}{N-1}
\sum_{p\in \mathbf{P}}
\left(\mathbf{P}(p)-\mu_\mathbf{P}\right)^2
},
\end{equation}
where \(\mathbf{P}\) denotes the luminance channel of the patch containing \(N\) pixels. $p$ is the coordinates of each pixel. $\mu_\mathbf{P}$ is the mean value of the luminance channel.
We further calculate the mean value of the
Laplacian-variance pyramid $\sigma_{lap}$~\cite{burt1983laplacian}:
\begin{equation}
\sigma_{lap} =
\frac{1}{L}
\sum_{l=0}^{L-1}
\mathcal{L}_l(\mathbf{P}),
\end{equation}
where $L$ is the number of pyramid levels. $\mathcal{L}_l$ denotes the Laplacian-variance at different filter scales. Patch groups whose
input patches lack texture (\ie, $\sigma_{img} < \tau_{img}$ or $\sigma_{lap} < \tau_{lap}$) are discarded. However, these statistics measure texture richness rather than moir\'e complexity. Some
moir\'e-free patches with rich textures may also be retained. 
To identify such patches, we process each remaining input patch with a pre-trained ESDNet and compute the SSIM between its output and the original input patch. Although ESDNet may not completely remove complex moir\'e patterns, it can still modify them while largely preserving clean patches. We therefore classify the input patch with an SSIM greater than $\tau_{ssim}$ as a clean patch with rich textures.
\paragraph{Optimal GT Selection.}
Each retained patch group comprises 1 input patch \(P_i\) and 5 candidate GT patches \(\{Q_i^k\}_{k=1}^{5}\). We select the highest-quality candidate GT patch as the final ground truth. for the \(i\)-th patch group, we feed each pair \((P_i, Q_i^k)\) into A-FINE~\cite{chen2025afine} to obtain a score. The candidate GT patch with the lowest score is selected:
\begin{equation}
    k^* = \arg\min_{k \in \{1,\ldots,5\}} \mathcal{D}(P_i, Q_i^k),
\end{equation}
where \(\mathcal{D}\) denotes the A-FINE~\cite{chen2025afine}, and a lower value indicates that $Q_i^k$ is better. \(k^*\) is the index of the selected candidate GT patch. If the improvement between the input patch score and the optimal candidate GT patch score is below a threshold \(\tau\), \ie,
\begin{equation}
  \mathcal{D}(P_i, P_i) - \mathcal{D}(P_i, Q_i^{k^*}) < \tau,
\end{equation}
we consider all candidate GT patches in the \(i\)-th patch group to be low-quality and discard the entire patch group.

\subsection{Dataset Statistics}
With \(\tau=15\), WildMoir\'e contains 6,832 moir\'e-GT pairs at 1024\(\times\)1024 resolution.Table~\ref{tab:prior_counts} shows the selected supervisions are 62.73\% (4,286) from GPT-Image-2~\cite{openai2026gptimage2}, 17.97\% (1,228) from Nano-Banana-2~\cite{google2026nanobanana2}, 9.10\% (622) from FLUX.2~\cite{bfl2025flux2}, 5.96\% (407) from Qwen-Image-Edit~\cite{wu2025qwenimage} and 4.23\% (289) from SDXL~\cite{podell2023sdxl}.
Table~\ref{tab:datasets} shows the comparison between our WildMoir\'e and existing demoir\'eing datasets.

\begin{table}[t]
\centering
\scriptsize
\setlength{\tabcolsep}{3.0pt}
\renewcommand{\arraystretch}{1.08}

\resizebox{\columnwidth}{!}{%
\begin{tabular}{cc|ccccc}
\toprule
& Model
& GPT-Image-2
& Nano-Banana-2
& FLUX.2
& Qwen-Image-Edit
& SDXL
\\
\midrule
& Ratio (\%)
& 62.73
& 17.97
& 9.10
& 5.96
& 4.23
\\
\bottomrule
\end{tabular}%
}
\caption{Distribution of models for the final selected supervisions in WildMoir\'e.}
\label{tab:prior_counts}
\end{table}

\begin{table}[t]
\centering
\small
\resizebox{0.98\columnwidth}{!}{%
\begin{tabular}{cccrc}
\toprule
Dataset &  Environment & Supervision & Resolution  & Size \\
\midrule
UHDM &  Controlled & Digital source & $\sim$4328$\times$3248 & 4,271   \\
DCID &   Controlled & Digital source & $\sim$4096$\times$3072 & 7,176   \\
WildMoir\'e   & In the wild & Generation & 1024$\times$1024 & 6,832   \\
\bottomrule
\end{tabular}%
}
\caption{Datasets used in the experiment. Size of UHDM is counted after our alignment-quality filtering. }
\label{tab:datasets}
\end{table}

\section{Data Augmentation and Mixing}
\subsection{Scale Transformation Data Augmentation}
We observe that real-world demoir\'eing datasets often exhibit imbalanced distributions of moir\'e scales, which will limit the generalization of the demoir\'eing models trained on them. To address this issue, we suggest a scale transformation data augmentation strategy, which simply rescales input images during training to diversify the scales of moir\'e patterns. Specifically, during training, we randomly apply \(s\times\) upsampling to moir\'e-GT pairs, and use the rescaled moiré images as inputs and the corresponding rescaled GTs as supervisions.
As the diversity of moiré scales in the training data increases, the generalization ability of the demoiréing model is improved to handle moir\'e patterns at different scales.

\subsection{Multiple Dataset Mixing}
Compared with existing demoiréing datasets, WildMoir\'e contains images with more complex moir\'e patterns. To enable the demoiréing model to handle both such complex patterns and the conventional moir\'e patterns represented in existing datasets, we additionally incorporate UHDM~\cite{yu2022esdnet} and DCID~\cite{mei2025image} into the training set, thereby further improving its generalization ability. The dataset details are provided in Table~\ref{tab:datasets}.

\begin{table*}[t]
\centering
\scriptsize
\setlength{\tabcolsep}{4pt}
\renewcommand{\arraystretch}{1.08}
\begin{tabular*}{0.9\textwidth}{@{\extracolsep{\fill}} c cl ccc ccc@{}}
\toprule
& \multirow{2}{*}{Models} & \multirow{2}{*}{Training Datasets} & \multicolumn{3}{c}{Full-reference Fidelity Metrics} & \multicolumn{3}{c}{No-reference Perceptual Metrics} \\
\cmidrule(lr){4-6} \cmidrule(lr){7-9}
& & & PSNR \(\uparrow\) & SSIM \(\uparrow\) & LPIPS \(\downarrow\) & MUSIQ \(\uparrow\) & TOPIQ \(\uparrow\) & Q-Align \(\uparrow\) \\
\midrule
\multirow{2}{*}{ \makecell[c]{Closed \\ Source} }  & GPT-Image-2 & -- & 15.1965 & 0.5265 & \textbf{0.4636} & \textbf{65.4375} & \textbf{0.5831} & 4.3665 \\
& Nano-Banana-2 & -- & \textbf{16.8793} & \textbf{0.5679} & 0.4696 & 60.9621 & 0.5002 & \textbf{4.4743} \\
\midrule
\multirow{2}{*}{ \makecell[c]{Open \\ Source} } & \multirow{2}{*}{\makecell[c]{ESDNet\\ (CNN-based)}} & UHDM+DCID & 21.5002 & 0.7341 & 0.3471 & 38.1971 & 0.2770 & 3.6343 \\
& & UHDM+DCID+WildMoir\'e & \textbf{23.3714} & \textbf{0.7821} & \textbf{0.2614} & \textbf{44.2109} & \textbf{0.3413} & \textbf{3.9607} \\
\midrule
\multirow{2}{*}{ \makecell[c]{Open \\ Source} } & \multirow{2}{*}{\makecell[c]{SDXL\\ (Diffusion-UNet-based)}} & UHDM+DCID & 21.9034 & 0.7522 & 0.2951 & 41.4789 & 0.3203 & 3.8128 \\
&  & UHDM+DCID+WildMoir\'e & \textbf{23.5644} & \textbf{0.7887} & \textbf{0.2621} & \textbf{45.0712} & \textbf{0.3583} & \textbf{4.0811} \\
\midrule
\multirow{2}{*}{ \makecell[c]{Open \\ Source} } & \multirow{2}{*}{\makecell[c]{Qwen-Image-Edit \\ (Diffusion-Transformer-based)}} & UHDM+DCID & 22.2314 & 0.7578 & 0.3101 & 55.3314 & 0.4276 & 4.1005 \\
& & UHDM+DCID+WildMoir\'e & \textbf{23.9313} & \textbf{0.7851} & \textbf{0.2528} & \textbf{58.9680} & \textbf{0.4624} & \textbf{4.2982} \\
\bottomrule
\end{tabular*}
\caption{Comparison of quantitative results. The optimal results in every part is highlighted in bold.}
\label{tab:main}
\end{table*}

\section{Experiments}
\subsection{Experimental Setup}
\paragraph{Evaluation Data.}
To evaluate demoir\'eing performance, we collected a separate test set of 247 additional moir\'e images. We obtain their ground truth images through camera capture rather than image generation to ensure the authenticity. Specifically, after capturing a moir\'e image, we adjusted the camera distance and viewing angle to capture a clean image of the same scene. We then spatially and photometrically align the clean image to its corresponding moir\'e image and use the aligned clean image as the ground truth.
\paragraph{Metrics.} We evaluate content fidelity using the full-reference metrics PSNR, SSIM~\cite{wang2004ssim}, and LPIPS~\cite{zhang2018lpips}. PSNR measures pixel-level fidelity, SSIM assesses structural similarity, and LPIPS captures perceptual similarity. We additionally report MUSIQ~\cite{ke2021musiq}, TOPIQ~\cite{chen2024topiq}, and Q-Align~\cite{wu2024qalign} as no-reference image-quality metrics to assess perceptual
  restoration quality.

\paragraph{Compared Settings.}
We train ESDNet, Qwen-Image-Edit, and SDXL on the combined UHDM and DCID datasets without scale transformation data augmentation, and use the resulting models as baselines~\cite{yu2022esdnet,wu2025qwenimage,podell2023sdxl,mei2025image}. We then augment the training set with WildMoiré and train both without and with scale transformation data augmentation. GPT-Image-2~\cite{openai2026gptimage2} and Nano-Banana-2~\cite{google2026nanobanana2} are closed-source models that cannot be fine-tuned; we therefore evaluate their official direct-editing outputs.

\paragraph{Implementation Details.}
All models are implemented in PyTorch~\citep{paszke2019pytorch}. For ESDNet, we use a batch size of 8 and optimize the model for 150 epochs using Adam~\citep{kingma2015adam} with an initial learning rate of $2\times10^{-4}$, $\beta_1=0.9$, and $\beta_2=0.999$. The learning rate is adjusted using cosine annealing with periodic warm restarts~\citep{loshchilov2017sgdr}. SDXL is fine-tuned following the InstructPix2Pix training formulation~\citep{brooks2023instructpix2pix}, with a batch size of 4 and a learning rate of $5\times10^{-5}$ for 150 epochs. Qwen-Image-Edit-2511 is fine-tuned using DiffSynth-Studio with LoRA~\citep{hu2022lora} rank 8, a batch size of 4, and a learning rate of $1\times10^{-4}$ for 150 epochs. Both training and inference for ESDNet are conducted on an NVIDIA RTX A6000 GPU, while SDXL and Qwen-Image-Edit are trained and evaluated on an NVIDIA RTX PRO 6000 GPU. During the ground truth selection process, we set the $\tau_{img}$ and the $\tau_{lap}$ to 30, the $\tau_{ssim}$ to 0.9. During training, we set the upsampling scale $s$ in scale transformation data augmentation strategy to 4.

\subsection{Experimental Results}
\paragraph{Quantitative Results.}
Table~\ref{tab:main} shows that incorporating WildMoir\'e
consistently improves the overall fidelity and perceptual quality
of a compact CNN restoration network and two generative editors.
Compared with the corresponding baselines, the complete
training setting with WildMoir\'e achieves clear improvements
across all three architectures and the overall metric suite.
The consistency across distinct model families indicates that
WildMoir\'e provides broadly useful supervision for complex
moir\'e removal rather than benefiting a particular training
mechanism or model architecture. The independent contribution of
scale transformation augmentation is further analyzed in
Table~\ref{tab:scale_aug}.

The commercial editors occupy a different fidelity--perception operating point. Their strong no-reference scores reflect clean and visually appealing outputs, but the substantially weaker full-reference metrics expose content mismatch. Nano-Banana-2 generally preserves the displayed layout and local content more faithfully than GPT-Image-2, yet it can leave visible moir\'e in difficult cases. GPT-Image-2 removes interference more aggressively but may introduce severe content or structural inconsistency; its outputs can also suppress physical screen-capture characteristics and resemble clean digital source images. These observations motivate using large generators as offline supervision producers instead of unrestricted test-time restorers.

\begin{figure*}[t]
\centering

    \includegraphics[width=1\linewidth]{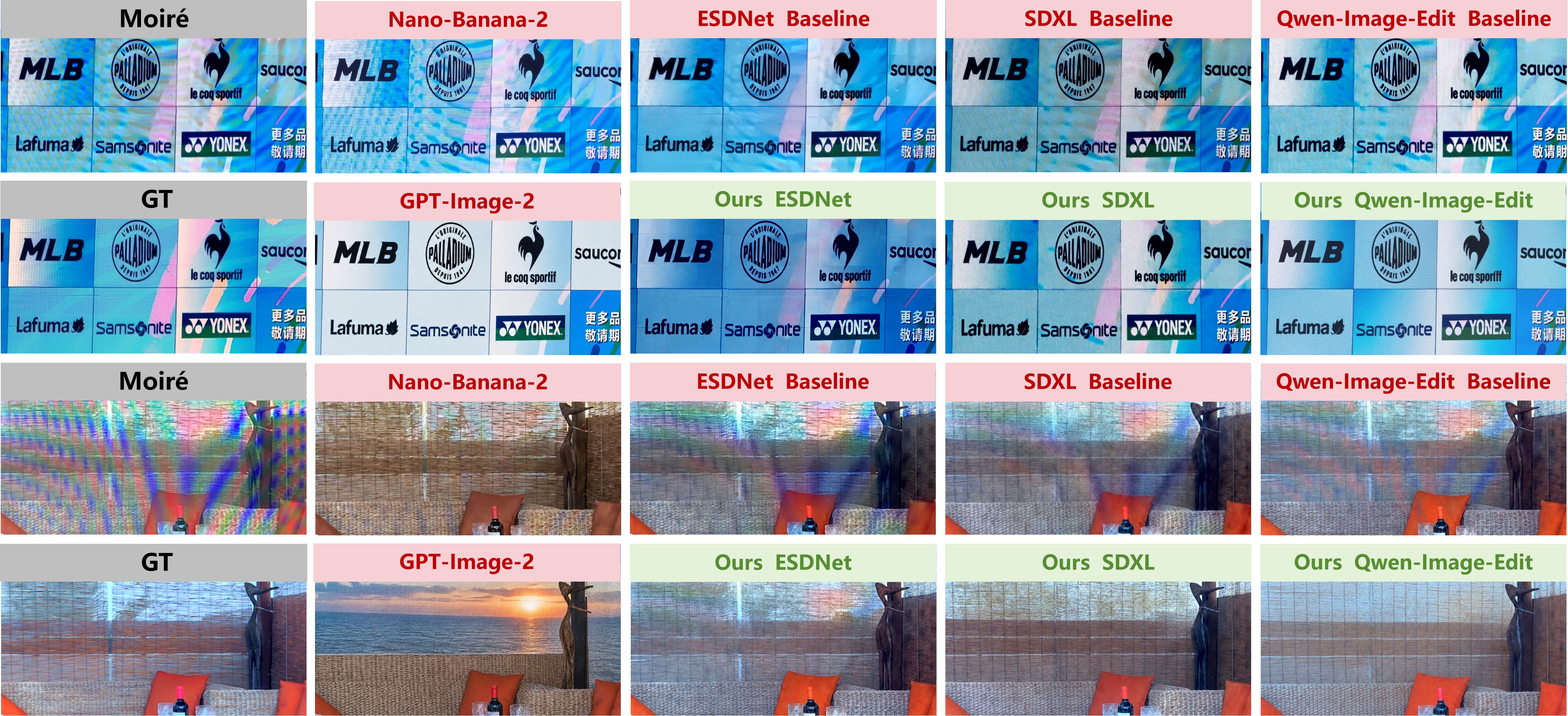}

\caption{Comparison of qualitative results.  The first example emphasizes logo and text fidelity under broad color interference, while the second contains severe mixed chromatic patterns over fine scene textures. Nano-Banana-2 is relatively content-faithful but may retain moir\'e, whereas GPT-Image-2 removes artifacts aggressively while replacing scene content and producing an overly digital-image appearance. Using WildMoir\'e suppresses residual bands more effectively across all 3 trainable models. }
\label{fig:qualitative}
\end{figure*}

\paragraph{Qualitative Results.}
Figure~\ref{fig:qualitative} supports the quantitative findings. The UHDM+DCID baseline models reduce part of the corruption but retain colored bands, local waves, or interference over fine texture. After adding WildMoir\'e, the same architectures recover cleaner logos, object boundaries, wall patterns, and furniture while retaining the photographic appearance of the display. The contrast with the two direct commercial editors also illustrates why perceptual quality alone is insufficient: an output can look clean while no longer representing the captured scene faithfully.

\subsection{Ablation Study}

Unless otherwise specified, all ablation experiments are conducted using ESDNet.

\paragraph{Effect of Candidate GT Sources.}
A central hypothesis of our framework is that different generators provide complementary candidate supervisions. We construct variants that restrict the candidate GT source to Nano-Banana-2 or GPT-Image-2 and train ESDNet with the same source data, per-epoch WildMoir\'e sampling, and augmentation recipe. Table~\ref{tab:prior_ablation} shows that GPT-Image-2 is the stronger individual source, but the five-prior construction achieves the optimal result on every metric. The method does not average candidate outputs; instead, it selects a locally suitable supervision. Therefore, a prior that is weaker on average can still contribute useful supervision at locations where its balance between artifact removal and content preservation is preferable.

\paragraph{Effect of Scale Transformation Augmentation.}
Table~\ref{tab:scale_aug} isolates the effect of scale
transformation augmentation on ESDNet under two training-data
settings. When trained on UHDM and DCID, enabling the augmentation
consistently improves all six metrics, showing that the transformed
samples help the model accommodate variations in the apparent
spatial scale of moir\'e patterns. A similar trend is observed after
WildMoir\'e is introduced: scale augmentation improves PSNR, SSIM,
MUSIQ, TOPIQ, and Q-ALIGN. These results indicate that the augmentation
is beneficial both for existing paired data and for the mixed
training setting. Meanwhile, the substantial improvement obtained
by adding WildMoir\'e remains evident even without scale
augmentation, confirming that the proposed dataset and the
augmentation strategy provide complementary gains.

\begin{table}[t]
\centering
\scriptsize
\setlength{\tabcolsep}{3.3pt}
\renewcommand{\arraystretch}{1.06}
\begin{tabular}{ccccccc}
\toprule
GT Source & PSNR\(\uparrow\) & SSIM\(\uparrow\) & LPIPS\(\downarrow\) & MUSIQ\(\uparrow\) & TOPIQ\(\uparrow\) & Q-A.\(\uparrow\) \\
\midrule
Nano-Banana-2 & 22.5640 & 0.7647 & 0.2921 & 40.9345 & 0.3087 & 3.8122 \\
GPT-Image-2 & 23.0013 & 0.7721 & 0.2742 & 42.0071 & 0.3236 & 3.8834 \\
Optimal of 5 Models & \textbf{23.3714} & \textbf{0.7821} & \textbf{0.2614} & \textbf{44.2109} & \textbf{0.3413} & \textbf{3.9607} \\
\bottomrule
\end{tabular}
\caption{Effect of candidate GT sources.}
\label{tab:prior_ablation}
\end{table}

\begin{table}[t]
\centering
\scriptsize
\setlength{\tabcolsep}{2.6pt}
\renewcommand{\arraystretch}{1.08}

\resizebox{\columnwidth}{!}{%
\begin{tabular}{ccccccccc}
\toprule
& Training Datasets
& Scale Aug.
& PSNR $\uparrow$
& SSIM $\uparrow$
& LPIPS $\downarrow$
& MUSIQ $\uparrow$
& TOPIQ $\uparrow$
& Q-ALIGN $\uparrow$
\\
\midrule

& UHDM+DCID
& \nomark
& 21.5002
& 0.7341
& 0.3471
& 38.1971
& 0.2770
& 3.6343
\\

& UHDM+DCID
& \yesmark
& \textbf{21.6414}
& \textbf{0.7468}
& \textbf{0.3230}
& \textbf{39.3647}
& \textbf{0.2941}
& \textbf{3.7478}
\\

\midrule

& UHDM+DCID+WildMoir\'e
& \nomark
& 23.3164
& 0.7710
& \textbf{0.2611}
& 43.1129
& 0.3224
& 3.9091
\\

& UHDM+DCID+WildMoir\'e
& \yesmark
& \textbf{23.3714}
& \textbf{0.7821}
& 0.2614
& \textbf{44.2109}
& \textbf{0.3413}
& \textbf{3.9607}
\\

\bottomrule
\end{tabular}%
}
\caption{Effect of scale transformation augmentation.
}
\label{tab:scale_aug}
\end{table}

\paragraph{Effect of Threshold for GT Selection.}
The A-FINE improvement threshold controls which local winners are sufficiently better than the imperfect input to serve as training supervisions. Table~\ref{tab:threshold} evaluates \(\tau\in\{5,10,15,20,25\}\) with ESDNet. Increasing the threshold from 5 to 15 progressively improves all six metrics while reducing the set from 8,991 to 6,832 pairs, showing that low-threshold candidates provide weaker or less consistent supervision. Performance declines when the threshold becomes more restrictive. At \(\tau=25\), only 2,818 pairs remain and the loss of content, degradation, and prior diversity outweighs the higher confidence of individual pairs. We therefore adopt \(\tau=15\), the optimal downstream operating point, to construct the final WildMoir\'e training set used throughout the paper.

\begin{table}[t]
\centering
\scriptsize
\setlength{\tabcolsep}{3.0pt}
\renewcommand{\arraystretch}{1.06}
\begin{tabular}{cccccccc}
\toprule
\(\tau\) & \#Pairs & PSNR\(\uparrow\) & SSIM\(\uparrow\) & LPIPS\(\downarrow\) & MUSIQ\(\uparrow\) & TOPIQ\(\uparrow\) & Q-A.\(\uparrow\) \\
\midrule
5 & 8,991 & 23.2863 & 0.7750 & 0.2720 & 43.0117 & 0.3309 & 3.8603 \\
10 & 8,143 & 23.3411 & 0.7807 & 0.2643 & 43.7824 & 0.3365 & 3.9431 \\
15 & 6,832 & \textbf{23.3714} & \textbf{0.7821} & \textbf{0.2614} & \textbf{44.2109} & \textbf{0.3413} & \textbf{3.9607} \\
20 & 5,284 & 23.3647 & 0.7786 & 0.2684 & 43.7306 & 0.3374 & 3.9513 \\
25 & 2,818 & 23.3021 & 0.7707 & 0.2925 & 42.9205 & 0.3315 & 3.8341 \\
\bottomrule
\end{tabular}
\caption{Sensitivity to the threshold for GT Selection.}
\label{tab:threshold}
\end{table}

\section{Conclusion}
We presented a multi-prior framework for constructing reliable generative supervision from unpaired photographs containing complex screen moir\'e, and built WildMoir\'e with 6,832 training pairs. For evaluation, we additionally built an independent test set comprising 247 pairs with captured clean ground truth. The framework leverages multiple generators as complementary visual priors and applies spatial and color alignment, patch filtering, and local optimal-supervision selection to obtain reliable paired training data. Mixing WildMoir\'e with UHDM and DCID consistently improves ESDNet, Qwen-Image-Edit, and SDXL without modifying their deployed architectures or introducing additional inference-time cost. The comprehensive ablations further validate the effectiveness of the proposed components, demonstrating that generative supervision improves robustness to challenging complex moir\'e patterns across different model architectures. These results also confirm the practical value of combining complementary generative priors with quality-controlled supervision for complex image demoir\'eing.

\clearpage
\appendix

\twocolumn[
\begin{center}
    \vspace{14mm}
    {\LARGE \textbf{Improving Complex Moir\'e Removal with Generative Supervision\\(Supplementary Material)}}
    \vspace{13mm}
\end{center}
]

% =========================================================
% Supplementary numbering
% =========================================================
\renewcommand{\thesection}{\Alph{section}}
\renewcommand{\thetable}{\Alph{table}}
\renewcommand{\thefigure}{\Alph{figure}}
\renewcommand{\theequation}{\Alph{equation}}

\setcounter{section}{0}
\setcounter{figure}{0}
\setcounter{table}{0}
\setcounter{equation}{0}

\section{Reliability of Generative Supervision}

Complex moir\'e images collected from public displays or online resources often lack exact clean source frames. Directly treating an unrestricted generative output as ground truth is nevertheless unreliable because a generator may remove interference while modifying text, objects, colors, or local structures. WildMoir\'e therefore uses generative models as offline candidate producers rather than as unrestricted restorers.

The resulting supervision should be regarded as a quality-controlled approximation of the unavailable clean reference. This is addressed in three ways. First, multiple priors increase the chance that at least one candidate provides a favorable balance between artifact removal and content preservation. Second, spatial and color alignment, synchronized local filtering, and A-FINE gating prevent unqualified outputs from being admitted directly. Third, WildMoir\'e is mixed with reliable digital-source supervision from UHDM and DCID rather than replacing existing paired data. The real paired datasets continue to anchor content fidelity, while WildMoir\'e expands the degradation distribution toward complex real-world patterns.

The downstream evidence in the main paper provides an additional practical validation. Multi-prior supervision outperforms supervision restricted to either GPT-Image-2 or Nano-Banana-2, and improvements are observed across a CNN, a diffusion UNet, and a diffusion Transformer. These results do not imply that every selected patch is error-free, but they indicate that the quality-controlled set contains useful supervision that is not tied to a particular restoration architecture.

\section{Additional Details}
\label{sec:filtering_details}

\subsection{Prompts of Candidate GT Generation }
\label{sec:prompts}

We employ GPT-Image-2~\citep{openai2026gptimage2}, Nano-Banana-2~\citep{google2026nanobanana2}, FLUX.2~\citep{bfl2025flux2}, SDXL~\citep{podell2023sdxl}, and Qwen-Image-Edit~\citep{wu2025qwenimage} as complementary generative priors. Each prior produces one candidate restoration for each input image. Table~\ref{tab:prompts} lists the exact English instructions used during data construction. GPT-Image-2 is given stronger content-preservation constraints because of its relatively aggressive editing behavior. Nano-Banana-2 receives a more explicit description of abnormal stripes and color regions to encourage more complete artifact removal. The three open-source priors use the same concise instruction. All three open-source priors are adapted with LoRA~\citep{hu2022lora} before candidate generation.

\begin{table*}[t]
\centering
\caption{Prompts used for candidate GT generation.}
\label{tab:prompts}
\small
\begin{tabular}{p{0.18\linewidth} p{0.75\linewidth}}
\toprule
Generative Models & Prompt \\
\midrule
GPT-Image-2 &
Remove the moir\'e patterns from the image. Strictly keep all other content unchanged. Preserve the original structure, colors, and fine details exactly. Ensure that all text on the screen remains clear, accurate, and completely unmodified. Preserve the original aspect ratio. \\
\midrule
Nano-Banana-2 &
Remove the moir\'e patterns, together with any abnormal stripes, color bands, or color blocks caused by the moir\'e effect. Strictly keep all other content unchanged, including the original structure, colors, and fine details. Ensure that all text on the screen remains clear, accurate, and completely unmodified. Preserve the original aspect ratio. \\
\midrule
SDXL, FLUX.2, and Qwen-Image-Edit &
Remove only the moir\'e patterns from the image. Keep all other content unchanged, including the colors, text, and fine details. \\
\bottomrule
\end{tabular}
\end{table*}

\subsection{Details of A-FINE}
\label{sec:afine_detail}

Conventional full-reference image-quality metrics generally assume that the reference image has perfect quality and measure how closely a test image matches it. This assumption is unsuitable for our setting because the available reference is the real input patch itself, which contains moir\'e. A restored candidate may therefore have higher perceptual quality than its reference while preserving the underlying content. A-FINE relaxes the perfect-reference assumption by adaptively combining a fidelity term and a naturalness term~\citep{chen2025afine}. For an imperfect reference $x$ and an evaluated image $y$, its score is
\begin{align}
D(x,y) &= F_{\eta}(x,y)+\lambda(x,y)N_{\gamma}(y), \label{eq:afine_score}\\
\lambda(x,y) &= \exp\!\left(k\left[N_{\gamma}(x)-N_{\gamma}(y)\right]\right), \label{eq:afine_lambda}
\end{align}
where $F_{\eta}$ evaluates fidelity to the reference, $N_{\gamma}$ evaluates the naturalness of the evaluated image, and $k>0$. Lower values of $F_{\eta}$, $N_{\gamma}$, and $D$ indicate better predicted quality. The adaptive weight allows naturalness to contribute more strongly when the evaluated image $y$ is more natural than an imperfect reference $x$, while fidelity dominates when the reference is substantially more natural. Consequently, using the imperfect moir\'e input as reference enables A-FINE to evaluate candidate quality while retaining an explicit constraint on fidelity to the captured content.

We use the official unscaled A-FINE output $D(x,y)$. This output can take negative values when the evaluated image is predicted to be substantially better than a low-quality reference, and lower values remain better. A-FINE is asymmetric, so the moir\'e patch must be used as the reference and the candidate as the evaluated image. To remain consistent with the notation in the main paper, we define
\begin{equation}
\mathcal{D}(P,Q),
\label{eq:afine_notation}
\end{equation}
where $P$ is the input moir\'e patch and $Q$ is a candidate restoration. For the $i$-th patch group, candidate selection and quality gating are performed as
\begin{align}
s_i^k &= \mathcal{D}(Q_i^k,P_i), \qquad
k^{*}=\arg\min_{k\in\{1,\ldots,5\}}s_i^k, \label{eq:afine_select_sup}\\
\Delta_i &= \mathcal{D}(P_i,P_i)-s_i^{k^{*}}. \label{eq:afine_improvement_sup}
\end{align}
The pair is retained only when $\Delta_i\geq\tau$. Comparing the local winner with the self-comparison score prevents the framework from accepting a candidate merely because it is the least poor result among five unreliable outputs.

\section{More Experiments}

\subsection{Statistics of Data Filtering}
The candidate-generation stage yields synchronized patch groups, each containing one real moir\'e patch and five aligned candidate restorations. The image-domain and Laplacian-pyramid statistics are computed jointly at the first filtering stage. A patch group is removed when either statistic falls below its corresponding threshold, which excludes regions with insufficient image variation. The remaining groups are processed by the ESDNet-response filter and the A-FINE selection stage described in the main paper.

Table~\ref{tab:filtering_funnel} reports the complete filtering funnel. Texture-complexity filtering removes 1,837 groups, while the ESDNet-response filter removes a further 889 groups. The final A-FINE gate retains 6,832 groups at $\tau=15$, corresponding to 50.18\% of the initial patch groups. This substantial reduction reflects the conservative objective of retaining only candidates that provide a clear quality improvement over the imperfect moir\'e observation.

\begin{table}[t]
\centering
\caption{\small Statistics of data filtering on constructing WildMoir\'e.}
\label{tab:filtering_funnel}
\small
\begin{tabular}{lrr}
\toprule
Stage & Groups & Retained \\
\midrule
Initial synchronized groups & 13,615 & 100.00\% \\
$\sigma_{img}$ and $\sigma_{lap}$ filtering & 11,778 & 86.51\% \\
ESDNet-response filtering & 10,889 & 79.98\% \\
A-FINE filtering ($\tau=15$) & 6,832 & 50.18\% \\
\bottomrule
\end{tabular}
\end{table}

\begin{table}[t]
\centering
\caption{\textbf{Statistics of the optimal GT distribution under different A-FINE filtering thresholds.}
Total denotes the number of selected priors, while the model columns
report row-wise percentages.}
\label{tab:threshold_prior_counts}
\small
\setlength{\tabcolsep}{3pt}

\begin{tabular}{@{}ccccccc@{}}
\toprule
$\tau$
& Total
& \makecell{GPT-\\Image-2}
& \makecell{Nano-\\Banana-2}
& FLUX.2
& \makecell{Qwen-\\Image-Edit}
& SDXL
\\[-1pt]
& \textnormal{(Number)}
& \textnormal{(\%)}
& \textnormal{(\%)}
& \textnormal{(\%)}
& \textnormal{(\%)}
& \textnormal{(\%)}
\\
\midrule
5  & 8,991 & 53.05 & 19.04 & 10.92 & 9.16 & 7.82 \\
10 & 8,143 & 57.40 & 19.26 & 10.86 & 6.95 & 5.54 \\
15 & 6,832 & 62.73 & 17.97 & 9.10  & 5.96 & 4.23 \\
20 & 5,284 & 65.93 & 19.00 & 7.51  & 4.86 & 2.69 \\
25 & 2,818 & 60.15 & 24.59 & 8.59  & 4.15 & 2.52 \\
\bottomrule
\end{tabular}
\end{table}

\subsection {Statistics of the Optimal GT Distribution}

Since the A-FINE filtering has the highest filtering strength, we further demonstrated the optimal GT distribution under different A-FINE thresholds.
Table~\ref{tab:threshold_prior_counts} reports the number of selected patches contributed by each prior under different A-FINE thresholds. Increasing $\tau$ makes the quality-improvement requirement more restrictive and reduces the dataset from 8,991 pairs at $\tau=5$ to 2,818 pairs at $\tau=25$.

From $\tau=5$ to $\tau=20$, the share of GPT-Image-2 increases from 53.05\% to 65.93\%, while the combined share of Qwen-Image-Edit and SDXL decreases from 16.98\% to 7.55\%. At $\tau=25$, the GPT-Image-2 share decreases to 60.15\%, whereas Nano-Banana-2 increases to 24.59\%. Generally, stricter filtering concentrates the retained set around the two proprietary priors and suppresses less frequent contributions from the open-source models.

Together with the threshold ablation in the main paper, these statistics illustrate a quality--quantity--diversity trade-off. Moderate filtering removes weak or inconsistent candidates, but an overly restrictive threshold substantially reduces both data volume and prior diversity. The adopted setting $\tau=15$ retains contributions from all five priors while achieving the best overall downstream metric balance.

\subsection{Results on UHDM and DCID Datasets}
We additionally utilized the filtering methods mentioned in the main text, including the image-domain standard deviation and Laplacian-variance pyramid with a threshold of 60 consistent with the data construction pipeline to select 11 and 378 complex moir\'e image samples from the UHDM and DCID datasets, respectively, for evaluation. The results are shown in Table \ref{tab:complex_existing}. Quantitative results demonstrate that the introduction of WildMoir\'e effectively boosts both full-reference fidelity and no-reference perceptual quality for challenging moir\'e samples across all three network architectures. Most evaluation metrics achieve stable and consistent performance gains, with only negligible fluctuation in the LPIPS score for the Qwen-Image-Edit model. These findings verify that WildMoir\'e can complement the limited complex moiré patterns in the original UHDM and DCID training distributions, thereby substantially enhancing the robustness of restoration models across diverse challenging moir\'e cases.

\begin{table*}[t]
\centering
\caption{\textbf{Quantitative results on challenging cases from the UHDM and DCID test sets.} The better result within each model pair is highlighted in bold.}
\label{tab:complex_existing}
\scriptsize
\setlength{\tabcolsep}{4pt}
\renewcommand{\arraystretch}{1.08}
\begin{tabular*}{0.9\textwidth}{@{\extracolsep{\fill}} c cl ccc ccc@{}}
\toprule
& \multirow{2}{*}{Models} & \multirow{2}{*}{Training Datasets} & \multicolumn{3}{c}{Full-reference Fidelity Metrics} & \multicolumn{3}{c}{No-reference Perceptual Metrics} \\
\cmidrule(lr){4-6} \cmidrule(lr){7-9}
& & & PSNR \(\uparrow\) & SSIM \(\uparrow\) & LPIPS \(\downarrow\) & MUSIQ \(\uparrow\) & TOPIQ \(\uparrow\) & Q-Align \(\uparrow\) \\
\midrule
\multirow{2}{*}{ \makecell[c]{Closed \\ Source} }  & GPT-Image-2 & -- & 16.4847 & 0.6288 & 0.3597 & \textbf{47.6108} & \textbf{0.4396} & \textbf{4.4022} \\
& Nano-Banana-2 & -- & \textbf{19.5363} & \textbf{0.7320} & \textbf{0.2859} & 45.6903 & 0.4112 & 4.2289 \\
\midrule
\multirow{2}{*}{ \makecell[c]{Open \\ Source} } & \multirow{2}{*}{\makecell[c]{ESDNet\\ (CNN-based)}} & UHDM+DCID & 26.5372 & 0.8722 & 0.2483 & 34.1202 & 0.3006 & 3.9790 \\
& & UHDM+DCID+WildMoir\'e & \textbf{26.9181} & \textbf{0.8784} & \textbf{0.2441} & \textbf{34.9631} & \textbf{0.3109} & \textbf{4.0698} \\
\midrule
\multirow{2}{*}{ \makecell[c]{Open \\ Source} } & \multirow{2}{*}{\makecell[c]{SDXL\\ (Diffusion-UNet-based)}} & UHDM+DCID & 26.5779 & 0.8706 & 0.2467 & 36.5014 & 0.3082 & 4.0967 \\
&  & UHDM+DCID+WildMoir\'e & \textbf{26.8603} & \textbf{0.8746} & \textbf{0.2458} & \textbf{37.1075} & \textbf{0.3187} & \textbf{4.1343} \\
\midrule
\multirow{2}{*}{ \makecell[c]{Open \\ Source} } & \multirow{2}{*}{\makecell[c]{Qwen-Image-Edit \\ (Diffusion-Transformer-based)}} & UHDM+DCID & 26.9703 & 0.8801 & \textbf{0.2360} & 44.7053 & 0.4118 & 4.2721 \\
& & UHDM+DCID+WildMoir\'e & \textbf{27.3071} & \textbf{0.8870} & 0.2367 & \textbf{46.8816} & \textbf{0.4174} & \textbf{4.3152} \\
\bottomrule
\end{tabular*}
\end{table*}

\subsection{More Visual Comparisons}

\begin{figure*}[p]
\centering

\IfFileExists{figures/more_result.png}{
    \includegraphics[
        width=0.98\textwidth,
        height=0.92\textheight,
        keepaspectratio
    ]{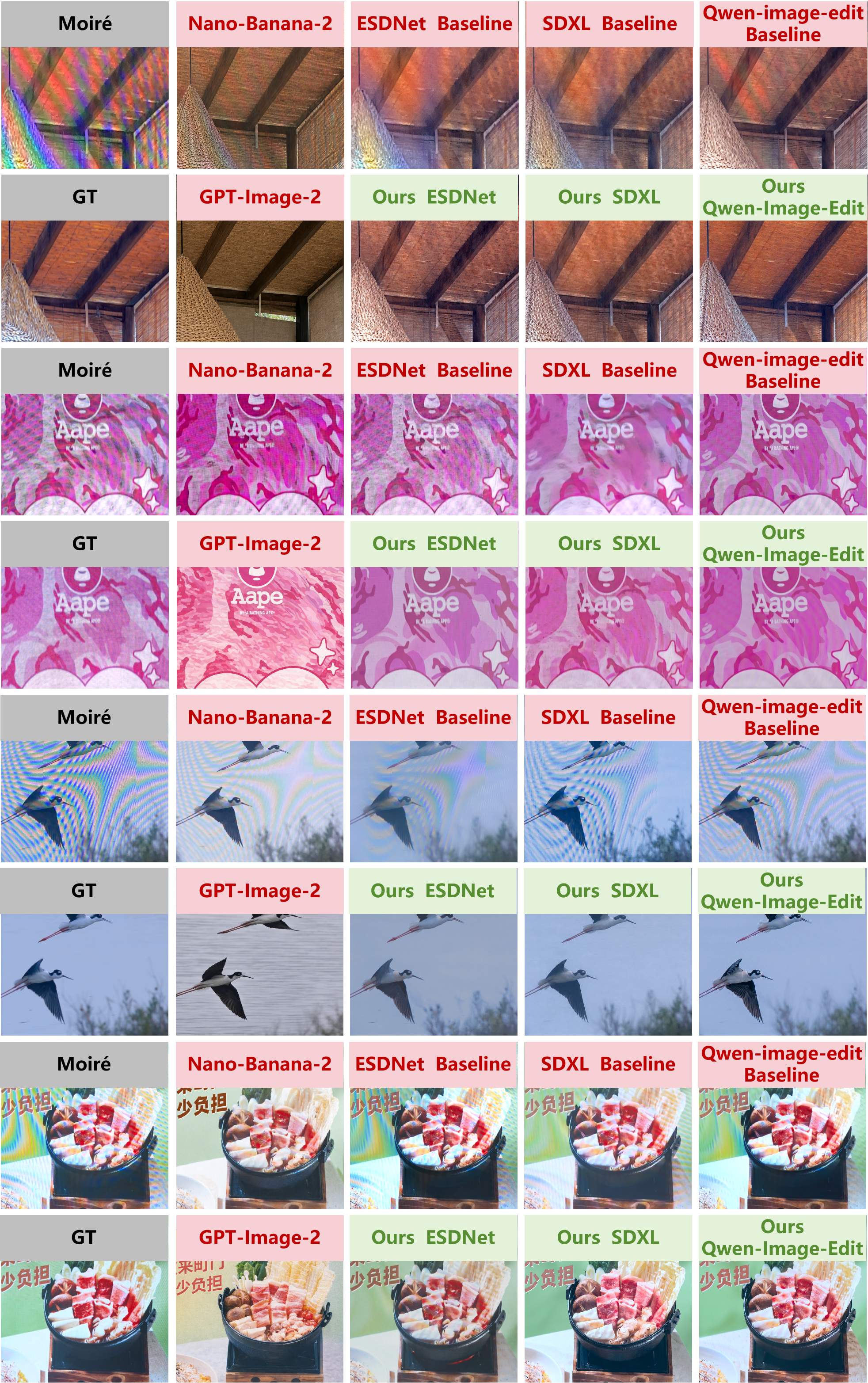}
}{
    \fbox{\parbox{0.90\textwidth}{
        \centering
        \rule{0pt}{0.65\textheight}
        \textbf{Placeholder for Additional Qualitative Comparisons on WildMoir\'e}
    }}
}

\vspace{-2mm}

\caption{
\textbf{Additional qualitative comparisons on our test set.}
Incorporating WildMoir\'e consistently improves complex moir\'e suppression
across ESDNet, SDXL, and Qwen-Image-Edit while preserving the captured content.
}
\label{fig:additional_visuals_a}

\end{figure*}

Figure~\ref{fig:additional_visuals_a} presents additional qualitative comparisons on the independently captured test set, which contains challenging complex moir\'e patterns such as broad chromatic bands, mixed-frequency interference, text-rich regions, and fine scene structures. Across ESDNet, SDXL, and Qwen-Image-Edit, models trained with WildMoir\'e consistently reduce residual colored bands and local wave patterns more effectively than their corresponding baselines, while preserving displayed text, object boundaries, and underlying image details.

\begin{figure*}[p]
\centering

\IfFileExists{figures/more_result_complex_subset.png}{
    \includegraphics[
        width=0.98\textwidth,
        height=0.92\textheight,
        keepaspectratio
    ]{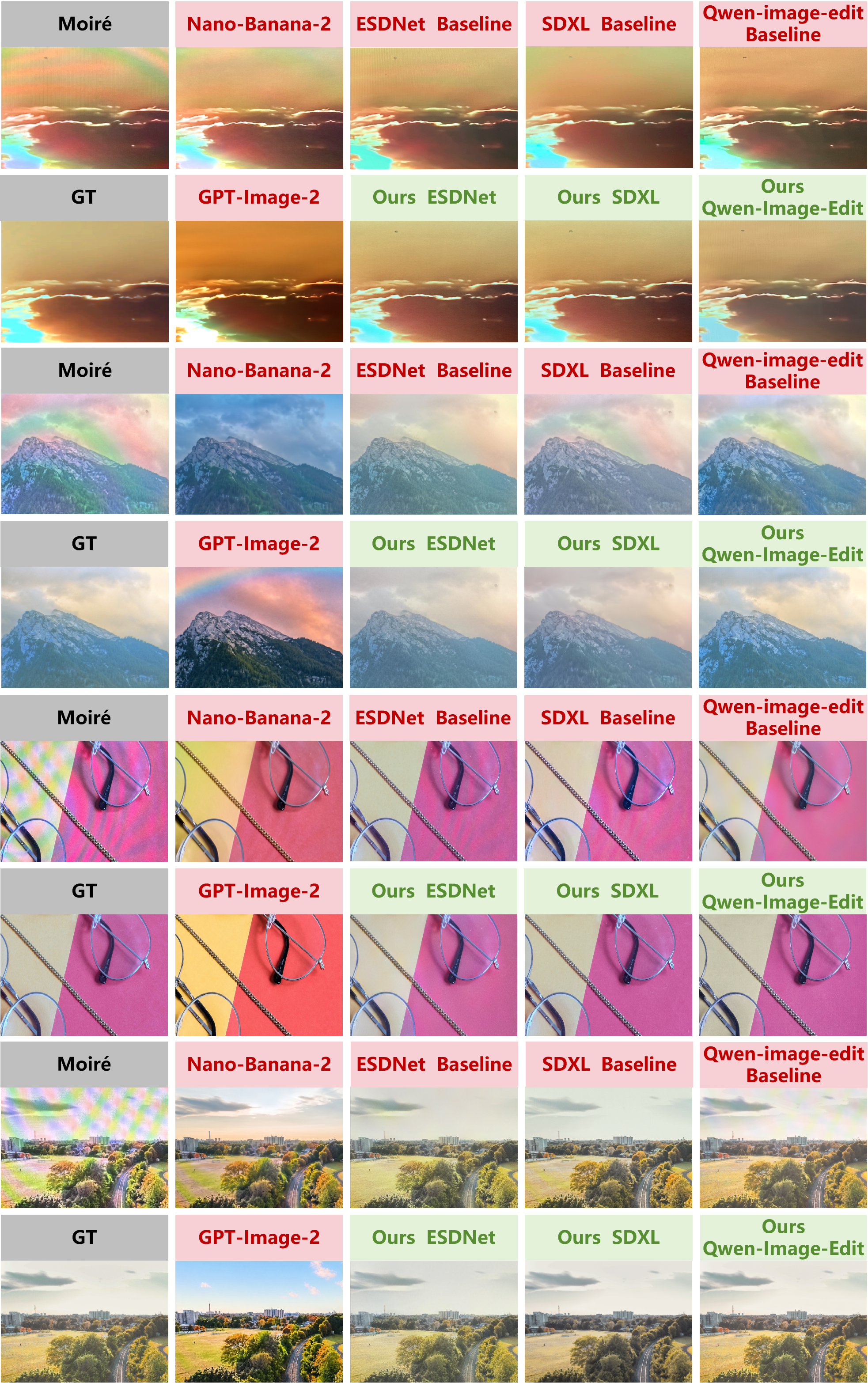}
}{
    \fbox{\parbox{0.90\textwidth}{
        \centering
        \rule{0pt}{0.65\textheight}
        \textbf{Placeholder for Additional Qualitative Comparisons on UHDM and DCID}
    }}
}

\vspace{-2mm}

\caption{
\textbf{Additional qualitative comparisons on challenging cases from the UHDM and DCID test sets.}
Models trained only on UHDM and DCID may retain broad color interference,
local waves, or mixed high- and low-frequency artifacts.
Incorporating WildMoir\'e improves complex moir\'e suppression across all
three trainable architectures while maintaining the captured content.
}
\label{fig:additional_visuals_b}

\end{figure*}

Figure~\ref{fig:additional_visuals_b} further presents challenging cases from the UHDM and DCID test sets. Although these samples are drawn from existing demoir\'eing benchmarks, they still contain complex moir\'e patterns that are difficult to remove using models trained only on the original UHDM and DCID data. Incorporating WildMoir\'e supervision improves restoration across all three architectures, indicating that the proposed generative supervision can complement existing training data and enhance robustness to challenging moir\'e patterns beyond our own captured test set.

\section{Limitations and Future Work}
\label{sec:limitations}

Our WildMoir\'e provides quality-controlled generative supervision for complex moir\'e images whose exact clean counterparts are unavailable, but it should not be regarded as a perfect substitute for digital-source ground truth. Although the proposed filtering and local selection procedures reduce unreliable candidates, subtle hallucinations, residual moir\'e, text modifications, or color inconsistencies may still remain. Complementary fidelity checks or human verification could further reduce metric-specific selection bias.

Future work will explore complementary quality-assessment models, manual or vision-language-model-assisted verification for ambiguous samples, and stronger constraints for preserving text and fine structures. Expanding the collection to more cameras, display types, venues, and shooting conditions may further improve degradation coverage.

\section{Use of Generative AI Tools}
Generative AI tools were used solely to assist with language polishing and were not used to generate figures, analyses, conclusions, or references. All content in the manuscript was carefully reviewed and verified by the authors.

\clearpage

\clearpage

\bibliography{AAAI2027}
\end{document}